# Educational robotics for children and their teachers[*]

Cristina Gena, Claudio Mattutino, Davide Cellie, and Enrico Mosca

Department of Computer Science,University of Turin, 10149, Italy
{name.surname}@unito.it

**Abstract.** This paper describes a Google Educator funded project devoted to the training of teachers (primary and secondary school) through an e-learning platform that will introduce them to educational robotics using Wolly, a social, educational and affective robot.

**Keywords:** educational robotics · communities of practice · computational thinking

## 1 Introduction

The term educational robotics refers to the use and development of learning environments based on robotic technology, mainly robots and software capable of programming them. Educational robots can play different roles, such as helping children to learn basic algorithms by programming the robots themselves, as well as their actions [1].
In our HCI lab, we carried out a co-design activity with children aimed at devising an educational robot called Wolly [2]. The main goal of the robot is acting as an affective peer for children: hence, it has to be able to execute a standard set of commands, compatible with those used in coding, but also to interact both verbally and affectively with students and, in the near future, adapt its behavior depending on the user, the context and the perceived user emotions. We designed the robot as open source project, made with a low-cost kit that could be easily reproduced and improved by anyone who wanted it.
In our vision, teachers and schools will be able to build one or more Wolly robots on their behalf, with help of their students, following our instructions, and through them carry out educational robotics activities guided by our mate- rial. However, as outlined by [6], there is the need, in all over the world, of big efforts aimed at growing school teachers with the appropriate skills to teach cod- ing and general principle of computer science. We believe that involving teachers in designing such courses could be a good solution for meeting their real educational needs and building communities of practice. Communities of Practice

[*] This work has been funded by the 2019 Google Educator PD Grant. We would like to thanks all the children and students that helped us in the co-design experiment. We would also thanks the anonymous reviewers of this paper that suggested us novel ideas and improvements to the research.

(CoP) are mutually supportive groups to assist newcomers as they find their way in a new endeavor [7]. Teachers could support one another at multiple levels: technical, pedagogical, practical.

Our ideas received the support from the Google Educator Grants Awards 2019 - EMEA[1] and thanks to this award we are carrying out the project with the help of our students, teachers, and young research fellows. The project, its goals and first implementations will be described in the followings.

## 2    The Wolly robot

The educational robotics activities will be proposed using the Wolly robot[2], designed and built by undergraduates and fellow students from our smart HCI lab, following a co-design approach, which involved primary school children in the conception and design phase, for more details see [2].

The first robot release is made of a very common hobby robotic kit, is able to move through its four independent motorized wheels, and can be controlled trough a web application. Its body has been almost completely 3D printed. The head of the robot is made of an Android-based smartphone able to show and understand emotion, and to produce vocal expression and to understand voice commend, for details see [3] .

The control logic of motorized components has been implemented in an Arduino Mega 2560 card, which interprets the commands to be executed via USB or serial to 115200 baud. It is connected to an Adafruit Motor Shield V1.2 motor driver, for communication with the 4 3-6 Volt DC motors, on which depends the wheel movement. A Servo MG995 of 5V DC and 400mA, tied to the head, allows small head rotations and helps to simulate an attentive behavior. A 6800mAh Powerbank LiPo battery with 5V 1A and 5V 500mA regulated outputs offers power to all the electronic cards. Finally a Wemos D1 mini card provides wireless communication with the robot, offers server functionalities and exposes REST APIs that provide a set of basic instructions (e.g., moveForward, turnLeft, turnRight, moveBackward, etc.) for controlling the motors and the servo. The head consists of a 1GHz quad core LG Smartphone, with 1.5GB of RAM and 16GB of ROM, running Android 7.0. The interaction with all the functions of movements, listening, speech and emotion management is implemented in a set of Android apps running on the smartphone and dialoguing also with external web services, and able to control the engines through the REST APIs, for details see [3].
The main goal of the robot is acting as an affective peer-tutor for children: it can execute a standard set of commands, compatible to the ones used while coding, but also to interact both verbally and affectively with students about their results, and in the future, it will adapt its behavior depending on the user's features, the context and the perceived user emotions. As long-term goal we plan to enrich the robot with a user modeling component, which will keep track of past interactions with the user, will reason about her based on her features, her skills, and her emotions recorded during past interactions [3]. In this way the robot will be able to adapt its interaction to the user and help her in a personalized way also considering her past preferences and choices [5]. As future work we plan to evaluate the children-robot interaction with real users considering the peculiarity of evaluations of both human-robot interaction [8] and of user-adaptive systems [4].

---
[1] https://edu.google.com/computer-science/educator-grants/index.html
[2] http://wolly.di.unito.it/